\documentclass[letterpaper, 10 pt, conference]{ieeeconf}  

\IEEEoverridecommandlockouts                              
\usepackage{graphicx}                       
\usepackage{graphics}                       
\usepackage{epsfig}                         
\usepackage[tight,footnotesize]{subfigure}  
\usepackage{booktabs}
\usepackage{amssymb,amsmath}
\usepackage{mdwmath}
\usepackage{mdwtab}
\usepackage{commath}                        
\usepackage{eqparbox}
\usepackage{bm,array}
\usepackage{multirow}
\usepackage[table,xcdraw]{xcolor}

\usepackage{stfloats}                       

\usepackage{hyperref}

\usepackage{cite}                           

\usepackage[T1]{fontenc}

\usepackage{tabularx}
\usepackage[table]{xcolor}

\title{\LARGE \bf
A 3D Object Detection and Pose Estimation Pipeline Using RGB-D Images
}

\author{Ruotao He, Juan Rojas and Yisheng Guan$^{1}$.
\thanks{$^{1}$Ruotao He, Juan Rojas and Yisheng Guan, the corresponding author (Email: ysguan@gdut.edu.cn) are with the Biomimetic and Intelligent Robotics Lab (BIRL), School of Electro-Mechanical Engineering, Guangdong University of Technology, Guangzhou, China, 510006.
        }%
}

\begin{document}

\maketitle
\thispagestyle{empty}
\pagestyle{empty}

\begin{abstract}
3D object detection and pose estimation has been studied extensively in recent decades for its potential applications in robotics. However, there still remains challenges when we aim at detecting multiple objects while retaining low false positive rate in cluttered environments. This paper proposes a robust 3D object detection and pose estimation pipeline based on RGB-D images, which can detect multiple objects simultaneously while reducing false positives. Detection begins with template matching and yields a set of template matches. A clustering algorithm then groups templates of similar spatial location and produces multiple-object hypotheses. A scoring function evaluates the hypotheses using their associated templates and non-maximum suppression is adopted to remove duplicate results based on the scores. Finally, a combination of point cloud processing algorithms are used to compute objects' 3D poses. Existing object hypotheses are verified by computing the overlap between model and scene points. Experiments demonstrate that our approach provides competitive results comparable to the state-of-the-arts and can be applied to robot random bin-picking.  
\end{abstract}
\section{INTRODUCTION}\label{sec:Intro}
3D object detection and pose estimation are of great significance to robotics because they allow robots to localize the objects. This capability enables robots to autonomously perform manipulation tasks such as pick and place, parts assembly, amongst other. Environmental uncertainty and lack of structure still pose important challenges to accurate and efficient object detection and pose estimation algorithms. Many algorithms have been developed to tackle this problem. Local image descriptors, such as SIFT \cite{lowe2004distinctive} and SURF \cite{bay2008speeded} are often used to match key points between the scene and textured objects. 2D keypoints are then back-projected to 3D, where the object's 6-DOF pose is retrieved based on the 3D point-to-point correspondences. However, these descriptors fail to extract stable feature points from texture-less objects common in industrial environments. Recently, learning-based methods \cite{brachmann2014learning}, \cite{tejani2014latent} use forest-based voting schemes on image patches to detect objects and estimate 3D poses. The former regresses object coordinates and estimates pose by minimizing an energy function. The latter integrates LINEMOD \cite{hinterstoisser2011multimodal} template patches into random forests and jointly estimates objects' positions and orientations. 
\begin{figure}[tbh]
    \centering
        \includegraphics[scale=0.44]{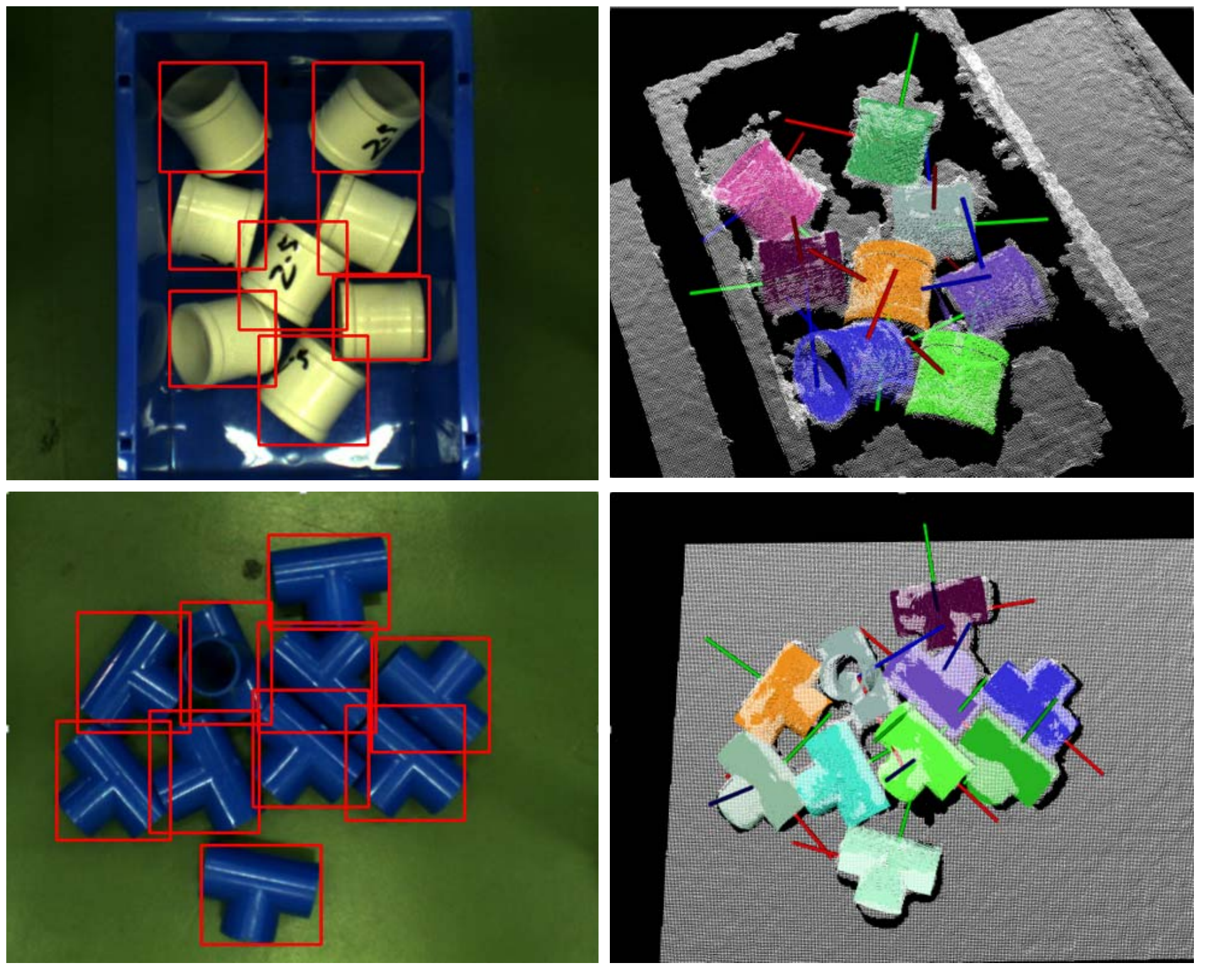}
        \caption{Left: Object detection results denoted with red bounding boxes Right: Pose estimation results highlighted with point clouds of different color and the corresponding coordinate axes.}
        \label{fig:demo of object detection}
\end{figure}
As more low-cost 3D cameras enter the market, the research focus has shifted to directly process 3D point clouds. The advantages of 3D-point-based algorithms are independence from object texture and invariance to illumination. The iterative closest point algorithm (ICP) \cite{ICP} is a common approach to align model points with scene points, but can only be used for pose estimation. However, an object's initial pose is crucial to the outcome accuracy. A few of the 3D descriptor algorithms \cite{spin,tombari2010unique,drost2010model,rusu2009fast,rusu2010fast,guo2013rotational} have been devised recently. The goal of 3D point cloud descriptors is to establish correspondences between model points and scene points. 6-DOF transformation can then be calculated based the 3D-point correspondences. In \cite{drost2010model}, the authors use point pair features for point matching and proposes a voting scheme to recover an object's 3D pose. In \cite{choi2012voting}, the point pair feature algorithm is enhanced by considering object boundaries. In \cite{guo2016comprehensive}, the authors compare the performance of popular 3D local descriptors on different datasets. These methods however still suffer from noise-sensitivity. They are prone to mismatch in cluttered scenarios or require rich variation in object geometry.

Template matching is another common approach for object detection. During offline training, an object's template images are sampled from varying viewpoints. During online testing, templates are compared to a scene image by computing the similarity. The object is detected if a template is matched. The object's pose is determined based on the template's training pose. In \cite{liu2012fast}, a fast directional chamfer matching performs edge-based template matching, which uses the chamfer distance and edge orientations as metrics \cite{chamfer}. In \cite{Cai2013Fast}, a two-stage cascaded detection method comprises of (i) a matching step that uses an index table to collect votes for templates and (ii) a verification step based on oriented chamfer matching \cite{Shotton2005Contour}.

LINEMOD \cite{hinterstoisser2011multimodal}, \cite{hinterstoisser2012gradient} combines silhouette gradient orientations from RGB images and surface normal orientations from depth images to represent object templates. Lookup tables are created \textit{a priori} for similarity indexing. Since template features are extracted from object boundary and depth image, LINEMOD is independent from object's texture and thus can detect texture-less objects. Later work \cite{hinterstoisser2012model} proposes a frame work of automatic template generation, object detection and pose estimation, adding color check and depth check for verification. 

\begin{figure}[tbh]
    \centering
        \includegraphics[scale=0.42]{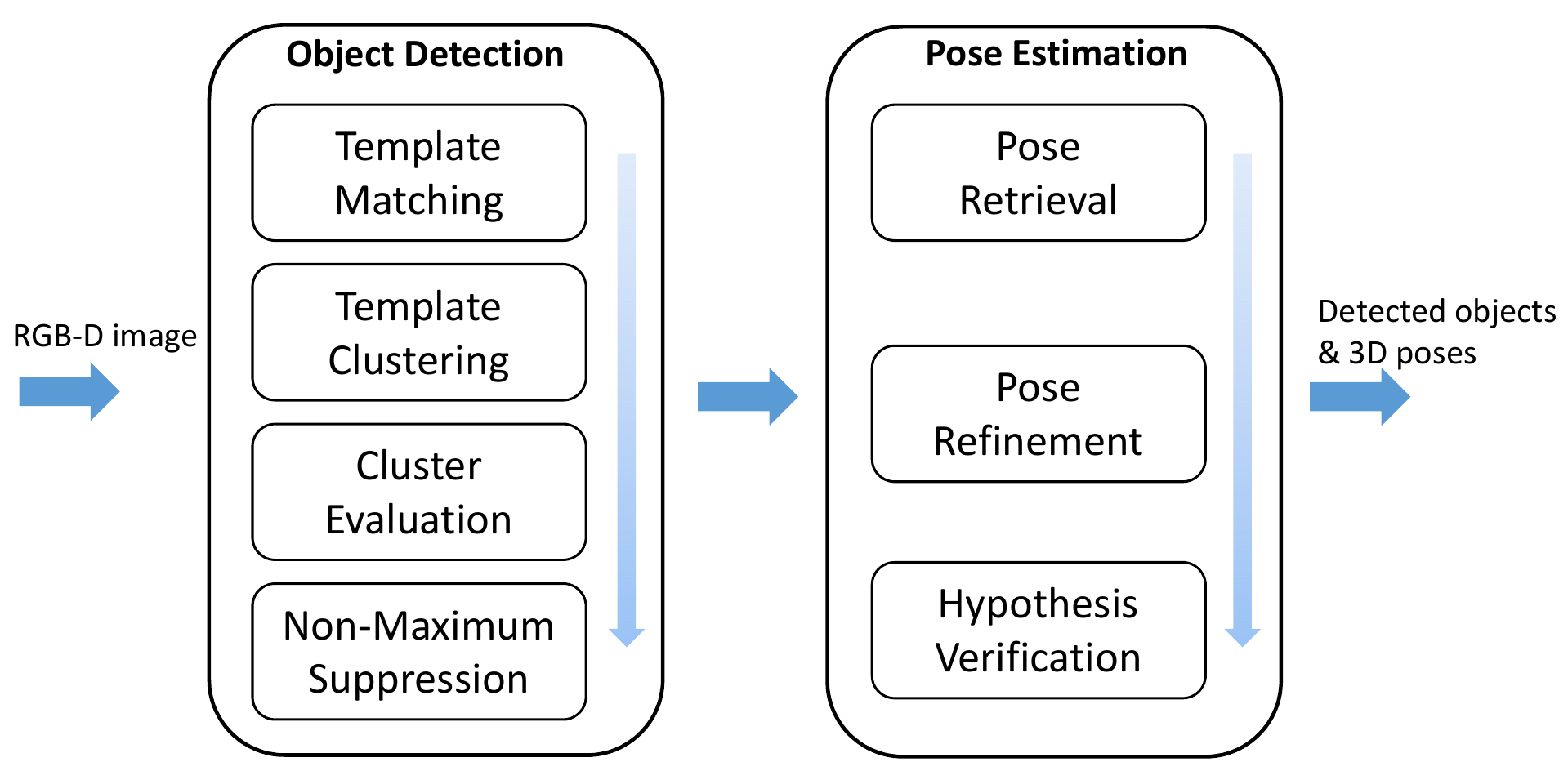}
        \caption{A proposed pipeline of 3D object detection and pose estimation.}
        \label{fig:pipeline}
\end{figure}

Like most of the template-based methods, LINEMOD need quantities of templates to fully cover an object. However, templates look alike if they have close training poses. Moreover, for objects of symmetric structure (cylinder, cone, etc.), template images from viewpoints distributed around the rotational axis will be identical. These ambiguous views will cause that multiple templates match with a single object. On the other hand, a large number of template will cause false positives in cluttered environment. The above issues are well illustrated in the images on the left column in Fig. \ref{fig:demo of dataset}. Since LINEMOD only produce the templates with a similarity score above an empirical threshold, it does not clearly output the number of instances of the same object. Obviously sorting the matches based on the similarity scores and only selecting the top results are not a solution to this problem because some matches could belong to the same target. Though \cite{hinterstoisser2012model} has improved LINEMOD in terms of choosing the appropriate number of training templates and lowering false-positive rate, it still does not provide a solution to directly detect multiple object instances. Our contribution in this paper is proposing a 3D object detection and pose estimation pipeline using RGB-D images, which extends LINEMOD to output multiple detected instances while reducing the false-positive rate. Given a single RGB-D image, our object detection starts with LINEMOD, producing initial matched templates. Then a simple but efficient template clustering step is performed, generating a set of template clusters, each of which represents an object hypothesis. False positives can be preliminarily filtered according to the number of templates in one cluster. Hypotheses are evaluated using their associated cluster of templates. Then, to remove duplicate hypotheses, we perform non-maximum suppression on images based on the evaluation score. At pose estimation scheme, we further use clustering to retrieve objects' initial poses from templates' training poses. Then, a specific sequences of point-cloud processing algorithms are used to refine object poses. Finally, a simple hypothesis verification step removes false positives by checking if object point cloud overlap well with the scene. Part of the results are presented in Fig. \ref{fig:demo of object detection}. The pipeline of our approach is illustrated in Fig. \ref{fig:pipeline} Experiments show that our approach, in terms of F1-Score, outperforms LINEMOD by 12.85\% and the method of Drost \textit{et al.} \cite{drost2010model} by 9.98\%. We also apply our approach to a robot random bin-picking task achieving an average success rate of 83.41\%.  

The rest of the paper is organized as follows: Sec. \ref{sec:object detection} presents object detection. Sec. \ref{sec:pose estimation} introduces 3D pose estimation. Sec. \ref{sec:experiment} demonstrates experimental results and Sec. \ref{sec:summary} summarizes key points in the paper.

\section{OBJECT DETECTION}\label{sec:object detection}
We present a template-matching object detection algorithm that consists of four phases: (i) LINEMOD is used to produce initial detection results. (ii) a template clustering algorithm is used to group templates with similar spatial location and refine the detection result. (iii) A scoring function is used to evaluate how template clusters coincide with the scene. (iv) Non-maximum suppression removes duplicate results based on the evaluation score.

\subsection{Template Matching} 
Our object detection starts with a template-matching algorithm, i.e., LINEMOD \cite{hinterstoisser2011multimodal}. LINEMOD defines a template as $T=(\{O_m\}_{m \in M}, P)$. $O$ is the template feature (gradient orientation or surface normal orientation). $M$ represents modalities (RGB image or depth image). $P$ is a list of feature locations $r$ in the template image. Then through a sliding-window approach, a template is compared to scene image $I$ at location $c$ based on a similarity measurement over its neighbors $R$:
$$\epsilon_s(I, T, c)=\sum_{r \in P}\max\limits_{t \in R(c+r)}f_m(O_m(r), I_m(t)). \eqno{(1)}$$
The function $f_m(O_m(r), I_m(t))$ measures cosine similarity for gradient orientations or surface normal orientations. A template is matched if the similarity score is higher than an empirical threshold. For more details on LINEMOD see \cite{hinterstoisser2011multimodal}. For our work, LINEMOD detected templates are used as an initial detection result. 

During training, template images are synthetically rendered. We need to fully cover an object with viewpoints from every angle. To do so, synthetic spheres are generated with the target object at the sphere's origin. Viewpoints are uniformly sampled on the sphere's surface. A virtual camera is then set on each viewpoint, taking images of object and object's poses $\{R, t\}$ w.r.t camera are saved for later use. To simulate different distances from object to camera, spheres with varying radii are also used.

\subsection{Template Clustering}
\begin{figure}[t]
    \centering
        \includegraphics[scale=0.40]{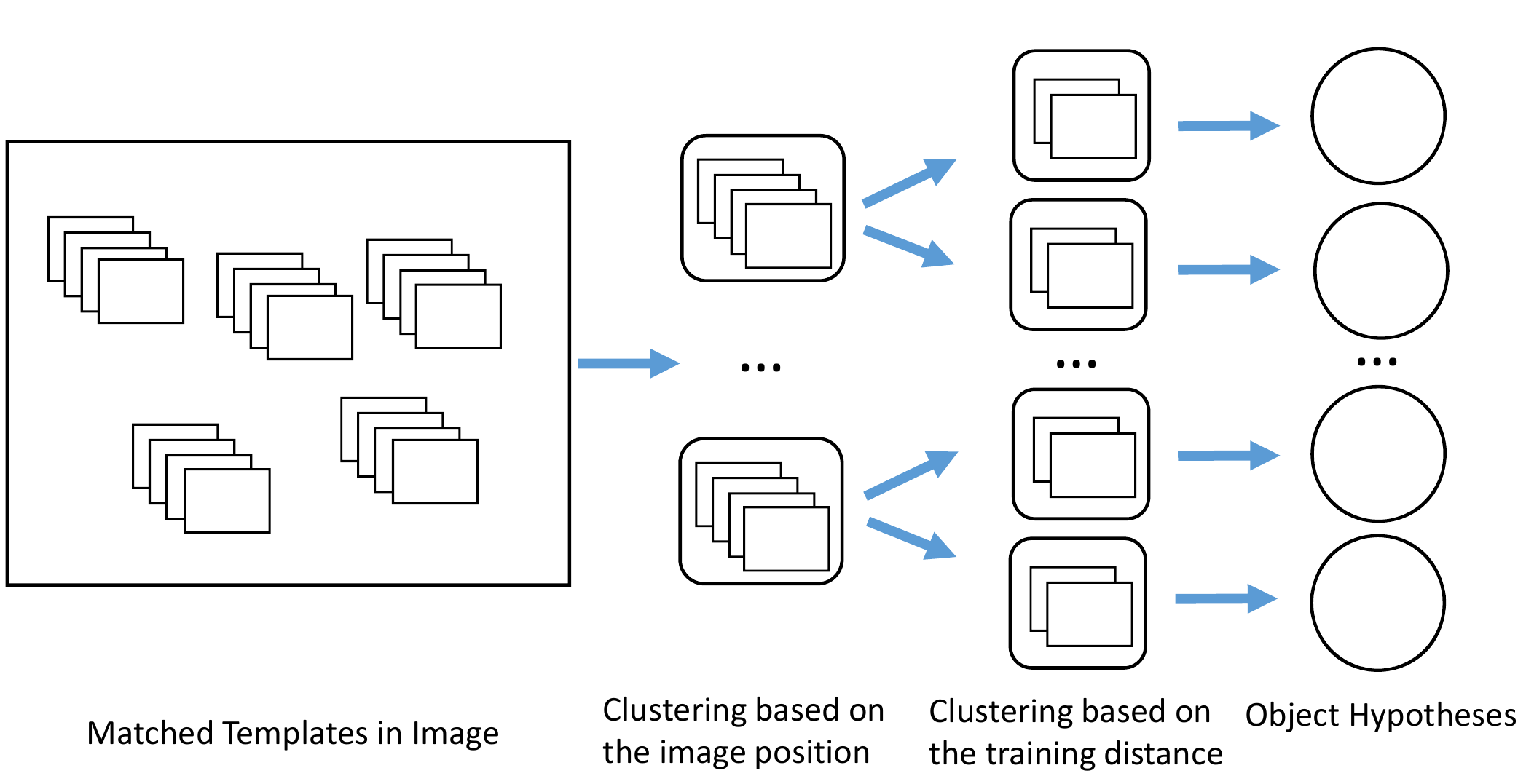}
        \caption{Multiple templates are matched in an image. Noticed that part of the templates are matched at approximate image positions due to objects' ambiguous views. We cluster the templates based on the matched positions $(r,c)$ and training distance $d$ and acquire object hypotheses.}
        \label{fig:template clustering}
\end{figure}

As is mentioned in Sec. \ref{sec:Intro}, LINEMOD detected templates could contain duplicate object instances and false positives. We thus use template clustering to aggregate templates with similar spatial location and preliminarily filter partial false positives. The templates are matched at position $(r, c)$ in the scene image, where $r$ and $c$ denote image row and image column respectively; while the distance $d$ between the object and the camera, was known \textit{a priori} during template training. Therefore, we define a matched template's spatial location as $F=(r, c, d)$. During the template clustering scheme, templates with similar spatial location are clustered together. The process consists of two steps. First, we quantize $r$ and $c$ with step $s_{im}$, so templates that have the same quantized position will be clustered into a group. Then, within each group, we further perform clustering that templates with the same $d$ will be assigned to the same group. Afterwards, each cluster contains a set of templates that have approximate matched positions and same training distances. We consider each cluster as an object hypothesis. Thus we average the position of all templates in each cluster to calculate object positions $(r_{obj}, c_{obj})$ in image as following shows:
$$r_{obj}=\frac{1}{n}\sum_{i=1}^{n}r_i. \eqno{(2)}$$ 
$$c_{obj}=\frac{1}{n}\sum_{i=1}^{n}c_i. \eqno{(3)}$$ Fig. \ref{fig:template clustering} illustrates the template clustering scheme.

One drawback of template matching is having a high false positive rate and LINEMOD is not excluded, especially for detecting objects of ambiguous shapes (cylinder, cuboid, etc.). We experimentally observe that clusters consisting of false positives normally have much fewer templates compared to clusters of correct matches. We can thus preliminarily filter part of false positive clusters based on templates count with a threshold $\tau_c$. $\tau_c$ is set according to the shape complexity of object that should be increased for objects with ambiguous shape and be decreased for objects with discriminated shape. Notice that the filter step is not expected to remove all the false positives and $\tau_c$ should be set loosely. The later hypothesis verification step introduced in Sec. \ref{sec:pose estimation} will further reject false positives.

\subsection{Cluster Evaluation}

For an object hypothesis, to measure the degree of similarity between a template cluster and its corresponding scene image, a similarity score $\gamma$ is defined according to Eqtn. 4:
$$\gamma=\alpha*\beta, \eqno{(4)}$$
where $\alpha$ is depth similarity and $\beta$ is normal similarity. $\alpha$ is computed by Eqtn. 5: 
$$\alpha=\frac{1}{e^{\frac{1}{m}\sum_{i=1}^m \epsilon_i}}, \eqno{(5)}$$
where $\epsilon$ is the sum of depth differences of the template image and the segmented scene image and $m$ is the number of matched templates in one cluster.
Likewise, $\beta$ is defined by Eqtn. 6:
$$\beta=\frac{1}{e^{\frac{1}{n}\sum_{i=1}^n \theta_i}}, \eqno{(6)}$$
 where $\theta$ stands for the sum of normal angle differences between the template image and the segmented scene image and $n$ is the number of matched templates in one cluster.
 
Aforementioned score calculation requires both depth images of templates and scene. To retrieve a template's depth image, we use its associated pose $\{R, t\}$ to render object model. To obtain the segmented scene depth image overlapped with the template, we create a 2D bounding box that has the same size as the template image, and use it to segment the scene image with its center at position $(r_{obj}, c_{obj})$. During calculation, we should notice that only a pair of valid depth values or a pair of valid surface normals will be taken into account. Each hypothesis will be assigned a score after each one is processed.

\subsection{Non-maximum Suppression}

A large quantity of object templates will produce plenty of object hypotheses, which happens more often in cluttered environment, e.g., a pile of objects in a bin. Moreover, some hypotheses might share the same image position $(r_{obj},c_{obj})$ but have different distances $d$ in our case. Thus these issues cause that multiple hypotheses belong to the same object. To address such problem, we perform non-maximum suppression on the image to remove duplicate results. For a hypothesis, a circle of radius $r$ is placed at $(r_{obj},c_{obj})$ and neighboring hypotheses are searched within the circle. We compare a hypothesis' score $\gamma$ to its neighbors' and if it is not the best among the neighbors, it will be suppressed. Only those hypotheses with locally maximum score will survive.

\section{Pose Estimation}\label{sec:pose estimation}

In this section, we present a 6-DOF pose estimation algorithm that consists of three parts: (i) an orientation clustering algorithm is used to retrieve object's initial pose; (ii) we use a specific sequence of point cloud processing algorithms for pose refinement; and (iii) we propose a term called collision rate to perform hypothesis verification.

\subsection{Initial Pose}

Since each object hypothesis contains a set of templates while templates are associated with training poses, we can fully utilize the training poses to compute an object's initial pose. We will first compute the orientation and then compute the position. For computing the orientation, simply averaging all orientations of template poses is not feasible since different views of one object could be similar while the poses could be rather different, which is especially true in symmetric objects. To solve this orientation averaging problem, we aggregate templates that have similar orientation together. To this end, we compute rotation between two poses using axis-angle representation $R(k,\theta_a)$. If $\theta_a$ is smaller than threshold $\tau_{\theta}$, we cluster the templates. After orientation clustering, we assume that cluster with most templates provides greatest confidence for object's orientation. We thus average all orientations inside the cluster. The result is considered as an object hypothesis's orientation.

For computing the position, we first use newly computed orientation to render object's CAD model to acquire model point clouds and the mask image. Then, the mask image is projected to the organized scene point clouds where scene points within the mask are segmented out. One would choose the centroid of the scene points as the object's origin, but the object model's origin will lie behind the surface point clouds due to self-occlusion. Such bias will probably lead subsequent ICP, which is very sensitive to initial pose, to a local optimum. To address this problem, we compute the translation vector from model point clouds to scene point clouds and apply it to the model's position that is known during rendering. The translation can be easily computed through subtracting the coordinates of a scene point from the coordinates of the corresponding model point. We compute both centroids of segmented scene points and model points to calculate the translation vector.

\subsection{Pose Refinement}

Due to sensor noises, scene points normally contain irregular points and missing holes while model points from CAD models are smooth. This inconsistency will the harm ICP registration. Therefore, we use moving least squares algorithm to reconstruct and smooth scene points \cite{alexa2003computing}. In addition, to ensure that the scene points and the model points have the same density, we apply a 3D voxel grid filter to sample them. One important parameter for ICP is the distance threshold $d_\tau$  for determining point correspondence. If the distance between a model point and its nearest scene point is above $d_\tau$, this pair of points will be ignored. We adopt a rough-to-fine ICP registration with varying $d_\tau$. At the rough stage, inspired by \cite{zeng2016multi}, $d_\tau$ is updated at each iteration with a high weight (90\%) of the largest distance of all point pairs until the algorithm converges or attains a maximum number of iterations. The model points will be transformed to an approximate pose. At the fine stage, $d_\tau$ is updated with a lower weight (45\%) that makes ICP ignore the noisy points. Then ICP can performs finer point clouds registration.

\subsection{Hypothesis Verification}  
To verify true-positive hypothesis and avoid false-positives, we introduce a term $\phi$ called collision rate for verification. We assume that transformed model points of a correct hypothesis will overlap well with the segmented scene points, i.e., model points will ``collide'' with scene points in large proportion. $\phi$ is defined as follows:
$$\phi=\frac{p_c}{p} \eqno{(7)}$$
where $p_c$ the number of model points that collide with scene points and $p$ is the total number of model points. To determine if a transformed model point collides with a scene point, we use an octree to partition the scene points by recursively subdividing them into eight octants until the volumes of octants meet a defined resolution. Then we check whether the model point falls within any of these octants. If $\phi$ is greater than predefined threshold $\tau_c$, the hypothesis is accepted.

\section{Experiments}\label{sec:experiment}
We quantitatively evaluate our method using the dataset of \cite{tejani2014latent}. The dataset contains 6 objects, each of which has a testing sequence consisting of over 700 images captured in near and far ranges. Compared to the dataset of Hinterstoisser \textit{et al}.\cite{hinterstoisser2012model} that only contains one object instance per image, the dataset of \cite{tejani2014latent} contains multiple object instances and foreground occlusions, which makes it more suitable and challenging for examining an algorithm's precision rate and recall rate. 

For each object, we render the template images from 216 viewpoints uniformly distributed on the synthetic sphere. Rotation invariance is achieved by rotating the camera around the optical axis from $-60^{\circ}$ to $60^{\circ}$ with a step of $10^{\circ}$. To achieve scale invariance, the camera is affixed to a set of spheres of 5 varying radii with a step of $0.1 m$. We thus generate 12960 templates per object.

In all tests, we use the metric proposed in \cite{hinterstoisser2012model} to determine if an estimation is correct. That is, for an object model $M$, given the estimated rotation $R_e$ and translation $T_e$, and the ground truth rotation $R_g$ and translation $T_g$, the matching score is defined as follows: $$m= \mathop{avg}\limits_{x\in M}||(R_gx+T_g)-(R_ex+T_e)|| \eqno{(7)}$$ The object is correctly detected and the pose is correctly estimated if $k_md>m$, where $k_m$ is a coefficient and d is the diameter of the object. For symmetric object that have a set of ambiguous views, the matching score is define as follows: $$m= \mathop{avg}\limits_{x_1\in M}\mathop{min}\limits_{x_2\in M}||(R_gx_1+T_g)-(R_ex_2+T_e)|| \eqno{(8)}$$ We set $k_m=0.15$ for all the tests.

\begin{figure*}[tbh]
    \centering
        \includegraphics[scale=0.7]{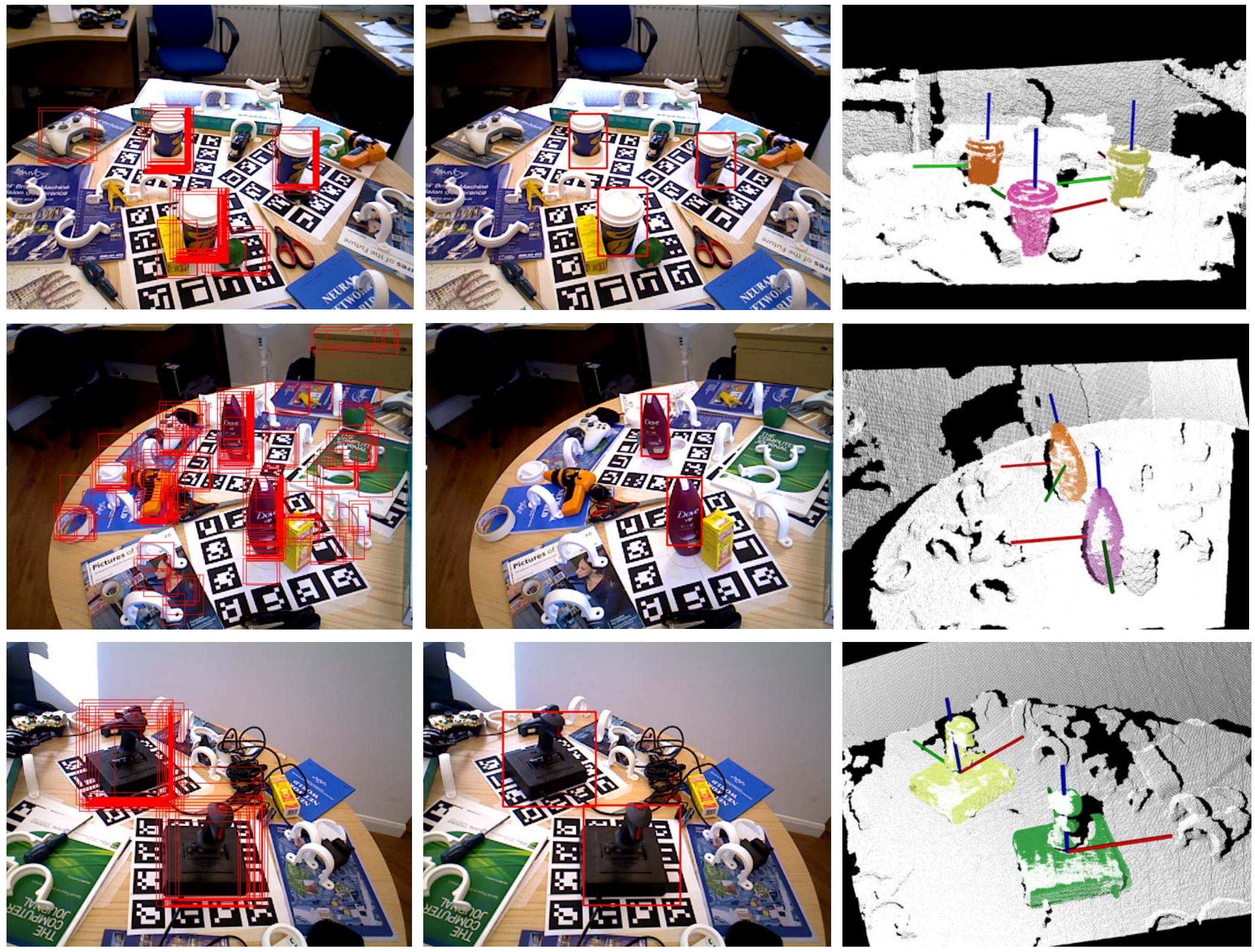}
        \caption{Snapshots of experiment results produced by our approach on the dataset of \cite{tejani2014latent} Left: Initial detection results produced by LINEMOD \cite{hinterstoisser2011multimodal} denoted with red bounding boxes Middle: Final detection results produced by the proposed approach. Right: Pose estimation results highlighted with point clouds of different color and the corresponding coordinate axes.}
        \label{fig:demo of dataset}
\end{figure*}

For evaluation, contrary to [20], we use F1-Score as the criterion, which is a weighted average of the precision and recall. The F1-Score is computed by $F1=2\frac{PR}{P+R}$, where $P$ is precision rate and $R$ is recall rate. We choose this type of evaluation because it can better reflect the algorithm's performance by considering both the precision and recall.

We compare our method to LINEMOD \cite{hinterstoisser2011multimodal}, the method of Drost et al.\cite{drost2010model} and Latent-Class Hough Forest (LCHF) \cite{tejani2014latent}. The result is listed in Table \ref{tb: f1}. We can see that our method achieve an average F1-score of 0.562 and outperforms LINEDMOD by 12.85\% and the method of Drost et al. by 9.98\%. LCHF achieve the highest score of 0.633. Fig. \ref{fig:demo of dataset} demonstrate part of the experimental results on the dataset \cite{tejani2014latent}. In Fig. \ref{fig:demo of dataset}, the images on the left column show the initial detection results produced by LINEMOD, denoted with red bounding boxes. The images on the middle column show final detection results and the right column displays the estimated poses highlighted with point clouds of different colors and the 3D axes.   

\begin{figure*}[t]
    \centering
        \includegraphics[scale=0.7]{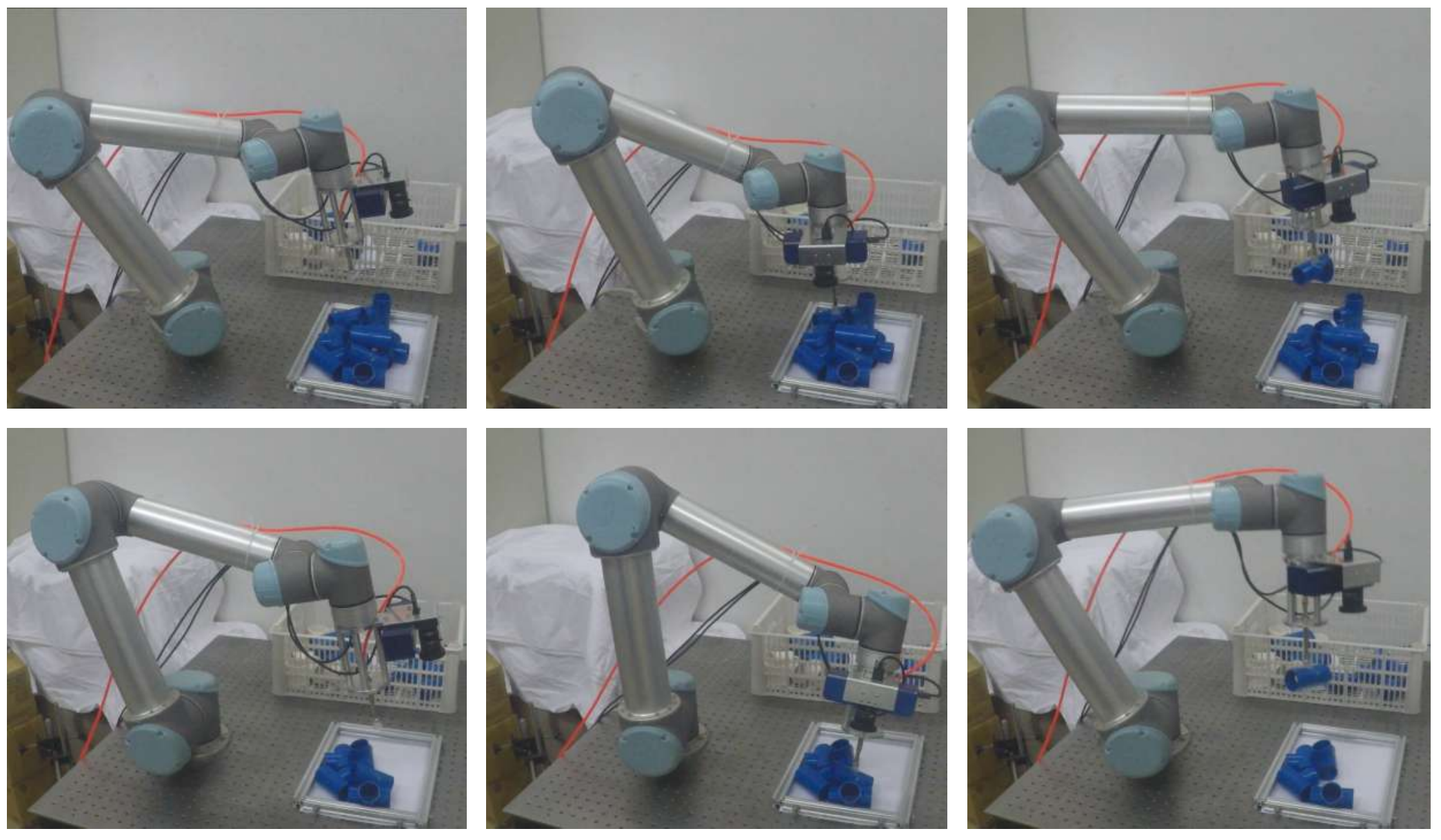}
        \caption{Snapshots of vision-guided bin-picking experiments using the proposed method. Each row demonstrates a standalone picking process. A pile of objects are randomly placed. The proposed method identifies objects and estimate the poses, guiding the robot to finish a pick-and-place task.}
        \label{fig:picking experiment}
\end{figure*}

We further verify our method by conducting bin-picking experiments. A 3D camera and an RGB camera are mounted on the wrist of the robot, providing the registered RGB-D images. The robot is equipped with a suction cup as an end-effector. As Fig. \ref{fig:picking experiment} shows, a pile of objects are randomly stacked together and the proposed method are adopted to detect the objects and estimate the poses. A grasping pose is generated according to objects' estimated pose. The robot performs pick-and-place manipulation based on the results. We select two kinds of objects for the experiments. By conducting 100 picking trials per object, we achieve an average success rate of 83.41\%. Notice that the simple design of the suction end-effector limits the feasibility of some grasping poses. For example, objects under certain poses expose not enough areas for suction, causing major picking failures. However, the solutions to the picking failures result from hardware design are out of the scope of this paper. Supplementary videos demonstrate part of the bin-picking process.


\newcolumntype{C}[1]{>{\centering\let\newline\\\arraybackslash\hspace{0pt}}m{#1}}

\begin{table}
\begin{center}
\scalebox{0.8}{
\begin{tabular}{|l|c|c|c|c|}
\hline
\textbf{Approach} & LINEMOD \cite{hinterstoisser2011multimodal} & Drost \textit{et al}.\cite{drost2010model} & LCHF \cite{tejani2014latent}& Our Approach \\
\hline
\hline
 \textbf{Sequences (\#pics)} & \textbf{F1-Score} & \textbf{F1-Score} &\textbf{F1-Score} &\textbf{F1-Score} \\
\hline
Coffee Cup (708) & 0.819 & 0.867 & \textbf{0.877} & 0.866 \\
\hline
Shampoo (1058)& 0.625 & 0.651 & 0.759 & \textbf{0.804} \\
\hline
Joystick (1032)& 0.454 & 0.277 & \textbf{0.534} & 0.512 \\
\hline
Camera (708)& 0.422 & 0.407 & 0.372 & \textbf{0.492} \\
\hline
Juice Carton (859)& 0.494 & 0.604 & \textbf{0.870} & 0.372 \\
\hline
Milk (860)& 0.176 & 0.259 & \textbf{0.385} & 0.323 \\
\hline
\textbf{Average (5229)} & 0.498 & 0.511 & \textbf{0.633} & 0.562 \\
\hline
\end{tabular}}
\end{center}
\caption{F1-Scores of LINEMOD \cite{hinterstoisser2011multimodal}, the method of Drost et al.\cite{drost2010model}, Latent-Class Hough Forest \cite{tejani2014latent} and our method for each object}
\label{tb: f1}
\end{table}

\section{CONCLUSIONS}\label{sec:summary}
We propose a 3D object detection and pose estimation pipeline using RGB-D images, which can detect multiple objects simultaneously while retaining low false positive rate. The approach starts with a template matching, generating a set of matches as initial results. Then, a clustering algorithm group the matched templates based on the spatial location, producing multiple object hypotheses. To remove duplicate results, a scoring function is utilized to evaluate the hypotheses and non-maximum suppression is performed. 3D pose of each object hypothesis is retrieved based on the training pose of the templates and further refined using a combination of point cloud process algorithms. Finally, hypotheses are verified by computing the overlap between the model points and scene points. Experiments on a public dataset show that our approach provides competitive results comparable to the-state-of-the-arts. We also apply the approach to a bin-picking context and proves the feasibility in a robotic application.

\addtolength{\textheight}{-12cm}   




\section*{ACKNOWLEDGMENT}
This work is supported by the grants: ``Major Project of the Guangdong Province Department for Science and Technology (2014B090919002), (2016B0911006).''





\bibliographystyle{ieeetr}
\bibliography{reference}

\end{document}